\documentclass[11pt, fleqn]{article}
\usepackage{preamble} 
\usepackage{macros}
\usepackage{letterfonts}
\usepackage{commands}

\title{The Lossy Horizon: Error-Bounded Predictive Coding for Lossy Text Compression (Episode I)}
\author{%
  Nnamdi Aghanya, Jun Li, Kewei Wang\\
  Aerospace, Transport and Manufacturing\\
  Cranfield University\\
  Cranfield, MK43 0AL \\
  \texttt{{nnamdi.aghanya, jun.li, kewei.wang.611} @cranfield.ac.uk} \\
}
\date{}
\begin{document}
\maketitle
\begin{abstract}
Large Language Models (LLMs) can achieve near-optimal lossless compression by acting as powerful probability models. We investigate their use in the lossy domain, where reconstruction fidelity is traded for higher compression ratios. This paper introduces Error-Bounded Predictive Coding (EPC), a lossy text codec that leverages a Masked Language Model (MLM) as a decompressor. Instead of storing a subset of original tokens, EPC allows the model to predict masked content and stores minimal, rank-based corrections only when the model's top prediction is incorrect. This creates a residual channel that offers continuous rate-distortion control. We compare EPC to a simpler Predictive Masking (PM) baseline and a transform-based Vector Quantisation with a Residual Patch (VQ+RE) approach. Through an evaluation that includes precise bit accounting and rate-distortion analysis, we demonstrate that EPC consistently dominates PM, offering superior fidelity at a significantly lower bit rate by more efficiently utilising the model's intrinsic knowledge.
\end{abstract}
%--------------------------------%
%            Document            %
%--------------------------------%

%%%%%%%%%%%%%%%%%%%%%%% DOCUMENT %%%%%%%%%%%%%%%%%%%%%%%%

\section{Introduction}
Lossy compression using LLMs is viewed as a viable option for the deployment of large language models post-training. Lossy compression can be achieved through the use of LLMs as decompression engines, where if a token is highly predictable from context, there is no need to store it~\cite{Deletang2024}.

Building upon this, we have identified and compared three families of lossy codecs. The first family of lossy codecs is Predictive Masking (PM), which serves as a baseline for our comparison and is based on the predictive masking of a token stream. The second family of lossy codecs is Error-Bounded Predictive Coding (EPC), which uses Error-Bounded Coding to provide a residual channel and a minimal amount of correction data stored in a rank index. The third family of lossy codecs is Vector Quantisation with a Residual Patch (VQ+RE), which is a baseline from the transform-coding paradigm. It provides vector quantisation of latent states while ensuring a bounded error in the final compressed output.

Our evaluations were conducted in terms of a rate (bits per character, BPC) versus distortion (fidelity) analysis for each codec. We also accounted for both the payload and static model costs for each codec, comparing the results to several well-known lossless compressors~\cite{Shannon1949, Witten1987}.

\section{Related Work}
Using predictive models for compression is a classical idea~\cite{Shannon1949, Rissanen1978}. Recent work leverages LLMs with arithmetic coding to approach the entropy limits for lossless text compression~\cite{Deletang2024, Valmeekam2023}. Our PM and EPC methods adapt this paradigm to a lossy setting, where an MLM provides a powerful conditional model for reconstruction. EPC's use of a rank-indexed residual stream is novel for text compression with LLMs, drawing inspiration from residual coding in other domains. VQ is a common technique for lossy compression of latent representations~\cite{Oord2017}; our VQ+RE baseline enhances it with an explicit error-correction layer, making it a competitive benchmark for text.

%%%%%%%%%%%%%%%%%%%%%%% DOCUMENT %%%%%%%%%%%%%%%%%%%%%%%%
\section{Methodology}\label{sec:method}
We standardise the models, datasets, and metrics in all experiments.

\paragraph{Models and Datasets}
Primary models are MLMs: \texttt{bert-base-cased} ("BERT-base"), \texttt{roberta-base} ("RoBERTa-base")~\cite{Liu2019}, and \texttt{distilroberta-base} ("DistilBERT")~\cite{Sanh2020}. The main corpus is \texttt{WikiText-103}. All datasets have non-overlapping train, validation, and test splits.

\paragraph{Metrics and Cost Accounting}\label{sec:metrics}
The rate is measured in BPC~\cite{Deletang2024}. Distortion is measured by character-level fidelity (1 - error rate), ChrF, and BERTScore~\cite{Zhang2019}. The amortised BPC accounts for the static size of the model, tokeniser, and any coders used over $N_{\text{copies}}$.

\paragraph{Entropy coding details (AC/rANS).}
We report the exact code lengths from our entropy coder: the Bernoulli flag stream and the $K$-way rank symbols are encoded using a standard range-ANS (rANS) with tabled frequencies. For the vocabulary-sized fallback token stream we report two quantities: (i) an \emph{ideal} adaptive arithmetic lower bound $\sum_{t}-\log_{2}\!\big((c_{x_t}+\alpha)/(C+\alpha V)\big)$ under an online unigram with Laplace smoothing ($\alpha{=}1$), and (ii) a practical rANS approximation in which, at each step, we encode a binary alphabet $\{s,\neg s\}$ with frequencies $\{c_s,\; C-c_s\}$ and then update counts (here $c_s$ is the current count of symbol $s$, $C$ the total count, and $V$ the vocabulary size). This $\{s,\neg s\}$ trick is a valid ANS coder and slightly overestimates the ideal bound because the complement mass is merged; implementing a full per-step $V$-way CDF would close this small gap but is orthogonal to our contribution and does not affect the relative RD trends.

\subsection{Compression Pipeline}
The pipeline consists of three stages: model specialisation, compression, and reconstruction.

\paragraph{Stage 1: Model Specialisation.}\label{p:s1}
To prepare the MLM for high-rate, predictability-based masking, we use an adaptive curriculum. The training data is split into a \texttt{fine-tuning\_set} (90\%) and a disjoint \texttt{policy\_set} (10\%). Each epoch, token predictability (surprisal) is computed on the \texttt{policy\_set} using the current model. This policy then dictates masking for the \texttt{fine-tuning\_set}, on which the model is updated. The masking rate increases linearly from 0.2 to 0.8 over several epochs to stabilise training~\cite{Weinshall2018, Kong2021}.

\paragraph{Stage 2: Compression Codecs.}
We evaluate three lossy codecs built upon the specialised MLM:

\begin{itemize}
    \item \label{subsec:pm_formal} \textbf{Predictive Masking (PM).} 
    Let $x_{1:N}$ be the token sequence. A subset of indices $\mathcal{M}$ are masked, while the remaining indices $\mathcal{S}$ are kept. We select $\mathcal{M}$ by identifying tokens with the lowest model surprisal, $s_i=-\log_2 q_\theta(x_i\mid X_{\setminus i})$, up to a target masking fraction $p_{\text{mask}}$. The payload consists of a bit vector for the positions of kept tokens and the kept tokens themselves, which are entropy-coded. Reconstruction involves deterministically infilling the masked positions $\{x_i\}_{i\in\mathcal{M}}$ by taking the most likely token, $\arg\max q_\theta(\cdot\mid X_{\mathcal{S}})$, from the specialised MLM (Figure~\ref{fig:pm_workflow}).
    
    \begin{minipage}{\dimexpr\linewidth-\leftmargin\relax}
        \centering
        \adjustbox{max width=\linewidth}{%
        \begin{tikzpicture}[
          font=\small\sffamily,
          node distance={12mm and 18mm},
          >=Latex,
          op/.style={
            rectangle,
            rounded corners=2pt,
            draw=black!60,
            fill=black!2,
            minimum height=9mm,
            minimum width=28mm,
            inner sep=3mm,
            align=center
          },
          arrow/.style={-Latex, line width=0.6pt},
        ]
        \node[op]                              (x)    {Input tokens\\$x_{1:N}$};
        \node[op, right=of x]                  (surp) {Compute surprisal $s_i$\\(MLM $q_\theta$)};
        \node[op, right=of surp]               (mask) {Select $\mathcal{M}$\\(low $s_i$)};
        \node[op, right=of mask]               (pos)  {Positions bitvector};
        \node[op, below=of mask]               (tok)  {Kept tokens $\mathcal{S}$\\entropy-coded};
        \node[op, right=of pos, minimum width=32mm] (mlm)  {MLM infill\\$\arg\max\, q_\theta(\cdot)$};
        \draw[arrow] (x)    -- (surp);
        \draw[arrow] (surp) -- (mask);
        \draw[arrow] (mask) -- (pos);
        \draw[arrow] (mask) -- (tok);
        \draw[arrow] (pos.east) -- (mlm.west);
        \draw[arrow] (tok.east) -| (mlm.south);
        \end{tikzpicture}
        }
        
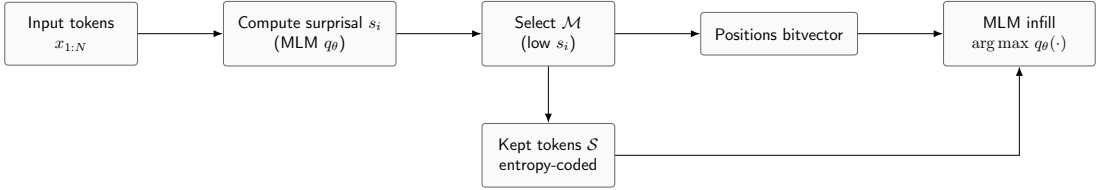
\captionof{figure}{Predictive Masking pipeline.}
        \label{fig:pm_workflow}
        \end{minipage}
    
    \item \label{subsec:epc_formal} \textbf{Error-Bounded Predictive Coding (EPC).}
    EPC begins identically to PM by selecting a mask set $\mathcal{M}$. However, instead of discarding the original tokens at masked positions, it introduces a residual stream. For each masked position $i \in \mathcal{M}$, if the model's top-1 prediction is incorrect, EPC stores a minimal correction. Let $r_i$ be the rank of the true token $x_i$ in the model's predictive distribution. If $r_i > 1$, EPC transmits an override flag followed by a compact representation of the correction. For ranks $2 \le r_i \le K$ (where $K$ is a hyperparameter), this correction is the rank index. For ranks $r_i > K$, EPC can optionally transmit the full token, allowing for lossless reconstruction of the masked subset. This design provides two controls: $p_{\text{mask}}$ trades kept tokens for model predictions, while the rank threshold $K$ controls the trade-off between the correction stream's bit rate and its error-bounding capability (Figure~\ref{fig:epc_workflow}).

    \begin{minipage}{\dimexpr\linewidth-\leftmargin\relax}
    \centering
    \adjustbox{max width=\linewidth}{%
    \begin{tikzpicture}[
      font=\small\sffamily,
      node distance={12mm and 18mm},
      >=Latex,
      op/.style={
        rectangle,
        rounded corners=2pt,
        draw=black!60,
        fill=black!2,
        minimum height=9mm,
        minimum width=30mm,
        inner sep=3mm,
        align=center
      },
      arrow/.style={-Latex, line width=0.6pt},
    ]
    \pgfdeclarelayer{background}
    \pgfsetlayers{background,main}
    \node[op] (x) {Input tokens\\$x_{1:N}$};
    \node[op, right=of x] (mask) {Select $\mathcal{M}$ by surprisal};
    \node[op, right=of mask] (flag) {Override flag $z_i$\\($r_i{>}1$?)};
    \node[op, right=of flag] (fb) {Fallback (if $r_i{>}K$)\\full token};
    \node[op, right=16mm of fb, minimum width=36mm] (mlm) {Decoder MLM\\apply overrides / $\arg\max$};
    \node[op, below=9mm of mask] (pos) {Positions bitvector};
    \begin{pgfonlayer}{background}
    \draw[arrow] (pos.east) -| ($(mlm.south)+(0,0mm)$);
    \end{pgfonlayer}
    \node[op, below=9mm of flag] (rank) {Rank index $r_i\!\in\![2,K]$\\(coded)};
    \draw[arrow] (x) -- (mask);
    \draw[arrow] (mask) -- (flag);
    \draw[arrow] (flag) -- (fb);
    \draw[arrow] (mask) -- (pos);
    \draw[arrow] (flag) -- (rank);
    \draw[arrow] (fb.east) -- (mlm.west);
    \draw[arrow] (rank.east) -| ($(mlm.south)+(0,0mm)$);
    \end{tikzpicture}
    }
    
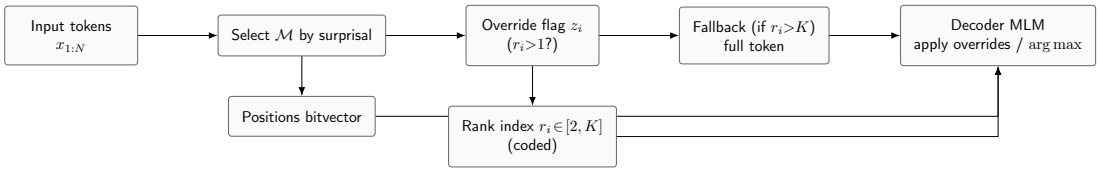
\captionof{figure}{Error-Bounded Predictive Coding pipeline.}
    \label{fig:epc_workflow}
    \end{minipage}
    
    \item\label{subsec:vq_formal} \textbf{Vector Quantisation with Residual Patching (VQ+RE)}
    As a transform-based baseline, we formalise a codec that operates in the model's latent space. This VQ pipeline compresses the key and value vectors within each attention head against learned codebooks. To mitigate train-test mismatch, the model is trained with quantised representations using scheduled self-feeding. The training objective combines the standard language modelling loss with VQ commitment and code utilisation losses. After an initial reconstruction, a residual patching step computes a diff between the original and reconstructed text. It transmits corrections for mismatched tokens using the same rank-based strategy as EPC, which guarantees a bounded error (Figure~\ref{fig:vq_workflow}).

    \begin{minipage}{\dimexpr\linewidth-\leftmargin\relax}
    \centering
    \adjustbox{max width=\linewidth}{%
    \begin{tikzpicture}[
      font=\small\sffamily,
      node distance={12mm and 18mm},
      >=Latex,
      op/.style={
        rectangle,
        rounded corners=2pt,
        draw=black!60,
        fill=black!2,
        minimum height=9mm,
        minimum width=30mm,
        inner sep=3mm,
        align=center
      },
      arrow/.style={-Latex, line width=0.6pt},
    ]
    \node[op]                              (x)    {Input tokens};
    \node[op, right=of x]                  (kv)   {Quantise K/V\\(EMA codebooks)};
    \node[op, right=of kv]                 (lm)   {LM forward\\with quantised states};
    \node[op, right=of lm]                 (rec)  {Initial decode};
    \node[op, below=9mm of rec]            (patch){Residual patch\\rank-based corrections};
    \node[op, right=16mm of rec, minimum width=32mm] (out) {Final text};
    \draw[arrow] (x)  -- (kv);
    \draw[arrow] (kv) -- (lm);
    \draw[arrow] (lm) -- (rec);
    \draw[arrow] (rec) -- (out);
    \draw[arrow] (rec) -- (patch);
    \draw[arrow] (patch.east) -| ($(out.west)+(0,-6mm)$);
    \end{tikzpicture}
        }
    
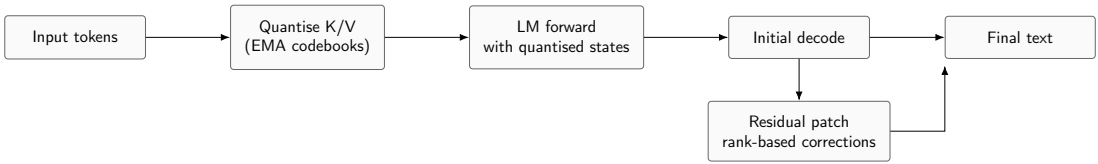
\captionof{figure}{VQ+RE pipeline.}
    \label{fig:vq_workflow}
    \end{minipage}
\end{itemize}
A complete derivation for each is in the \hyperref[asec:formalism]{Appendix}.

\paragraph{Stage 3: Reconstruction and Refinement.}
For all codecs, decompression involves deterministic, confidence-ordered infilling. For EPC and VQ+RE, this can be iterated. Let $g_\phi$ be a small refinement head and $\hat{X}^{(0)}$ be the initial decode. For $s=1,\dots,T_{\text{it}}$ (default $T_{\text{it}}=2$):
\begin{equation}
\label{eq:refine}
\begin{aligned}
u_i^{(s)} &= 1 - \max_v p_\phi(v \mid \hat{X}^{(s-1)}, i), \\
\mathcal{U}^{(s)} &= \text{top-}M_s \text{ indices of } u^{(s)}, \\
\hat{x}_i^{(s)} &= \arg\max_v p_\phi(v \mid \hat{X}^{(s-1)}, i), \ \forall i \in \mathcal{U}^{(s)}.
\end{aligned}
\end{equation}
This improves predictions without extra side information. The refinement head is counted in the static model size.

\subsection{Implementation and Evaluation Protocol}
Experiments are run on NVIDIA V100 GPUs using \texttt{PyTorch} and \texttt{Hugging Face}. The key results are averaged over five random seeds. We fine-tune the MLM as described in \hyperref[p:s1]{Stage 1}, then generate RD curves by sweeping key parameters for each codec: masking rate $p_\text{mask}$ for PM and EPC, and rank threshold $K$ for EPC and VQ+RE's residual stream. The curves plot BPC vs. Fidelity, providing a clear comparison of their efficiency.

%%%%%%%%%%%%%%%%%%%%%%% DOCUMENT %%%%%%%%%%%%%%%%%%%%%%%%
\section{Experiments and Results}\label{sec:results}

\paragraph{Protocol.}
We evaluate the three codecs from \S\ref{sec:method} on held-out text from the \texttt{WikiText-103} test split. Unless stated, each configuration is run over $5$ seeds. We sweep masking rates $p_{\text{mask}}\in\{0.2,0.4,0.6,0.8\}$, EPC rank thresholds $K\in\{4,16,64,128\}$ and (for VQ+RE) anchor stride $\in\{0,16,32,64\}$. Position streams use the "min" positional coder (the better of enumerative/RLE on each sequence). We record the rate in BPC and distortion via character-level fidelity ("CharFid"), ChrF, and BERTScore. All curves use the amortisation conventions from \S\ref{sec:metrics}; because each within-model comparison holds static artefacts fixed (MLM, tokeniser, coder tables), relative RD comparisons are unaffected.

\subsection{Main rate–distortion results (CharFid)}
\begin{figure}[ht]
  \begin{adjustbox}{width=0.92\linewidth, center}
    \includegraphics{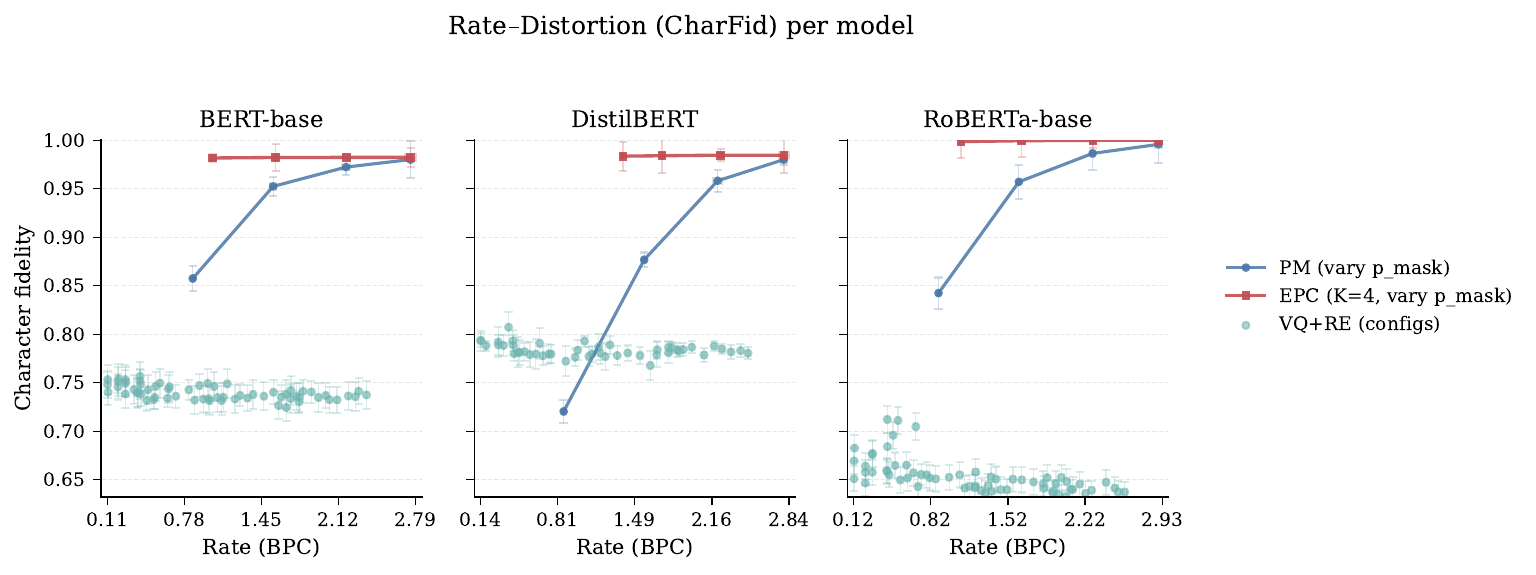}
  \end{adjustbox}
  \caption{\textbf{Rate–distortion (CharFid vs. BPC).} EPC consistently Pareto-dominates PM in the high-fidelity region. VQ+RE occupies a different RD regime, trading much lower rates for substantially lower fidelity. Error bars (where visible) show $\pm1$ sd over seeds; PM/EPC variances are negligible.}
  \label{fig:rd-main}
\end{figure}

Figure~\ref{fig:rd-main} plots CharFid vs. BPC for all three codecs and models. Three consistent trends emerge:

\begin{enumerate}
\item \textbf{EPC dominates PM in the high-fidelity regime.}  
Across models, to reach $\approx\!0.98$ CharFid, EPC reduces the bit rate by \textbf{52–63\%} relative to PM (Table~\ref{tab:at98}).
\begin{itemize}
\item \textbf{BERT-base:} PM needs $2.744$ BPC at 0.980 CharFid, while EPC achieves $0.982$ CharFid at $1.028$ BPC (\textbf{62.5\%} fewer bits).
\item \textbf{DistilBERT:} PM uses $2.792$ BPC at 0.980; EPC attains $0.982$ at $1.321$\,BPC (\textbf{52.7\%} fewer).
\item \textbf{RoBERTa-base:} PM uses $2.293$ BPC at 0.986; EPC attains $0.999$ at $1.097$\,BPC (\textbf{52.2\%} fewer).
\end{itemize}

\item \textbf{At a fixed rate, EPC yields much higher fidelity.}  
Around $1.6$\,BPC, EPC improves CharFid by \textbf{+2.9 to +10.7} absolute points over PM (Table~\ref{tab:at16bpc}). On DistilBERT, PM is $0.877$ at $1.571$ BPC, while EPC is $0.984$ at $1.727$ BPC (\textbf{+10.7} points). On BERT-base, EPC improves from $0.952$ to $0.981$ at a near-identical rate.

\item \textbf{Model strength matters.}  
RoBERTa-base provides the strongest RD curve: EPC retains $\geq\!0.995$ CharFid at $\approx\!1.10$\,BPC (Table~\ref{tab:at98}), while BERT-base and DistilBERT plateau near $\approx\!0.982$ CharFid.
\end{enumerate}

\begin{table}[ht]
\centering
\small
\caption{\textbf{Bits to reach $\approx$0.98 CharFid.} EPC reduces rate 52–63\% vs. PM at matched fidelity.}
\label{tab:at98}
\begin{tabular}{lcccccc}
\toprule
& \multicolumn{3}{c}{PM} & \multicolumn{3}{c}{EPC (best)} \\
\cmidrule(lr){2-4} \cmidrule(lr){5-7}
Model & BPC $\downarrow$ & CharFid $\uparrow$ & $p_{\text{mask}}$ & BPC $\downarrow$ & CharFid $\uparrow$ & $(p_{\text{mask}},K)$ \\
\midrule
BERT-base      & 2.744 & 0.980 & 0.2 & \textbf{1.028} & \textbf{0.982} & (0.8, 4) \\
DistilBERT     & 2.792 & 0.980 & 0.2 & \textbf{1.321} & \textbf{0.982} & (0.8, 16) \\
RoBERTa-base   & 2.293 & 0.986 & 0.4 & \textbf{1.097} & \textbf{0.999} & (0.8, 4) \\
\bottomrule
\end{tabular}
\end{table}

\begin{table}[ht]
\centering
\small
\caption{\textbf{Fidelity at $\approx$1.6\,BPC.} EPC improves CharFid by +2.9 to +10.7 points over PM.}
\label{tab:at16bpc}
\begin{tabular}{lcccc}
\toprule
Model & PM (BPC, CharFid) & EPC (BPC, CharFid) & $\Delta$BPC & $\Delta$CharFid \\
\midrule
BERT-base    & (1.552, 0.952) & (1.580, 0.981) & +0.028 & \textbf{+0.029} \\
DistilBERT   & (1.571, 0.877) & (1.727, 0.984) & +0.156 & \textbf{+0.107} \\
RoBERTa-base & (1.623, 0.957) & (1.650, 0.999) & +0.027 & \textbf{+0.042} \\
\bottomrule
\end{tabular}
\end{table}

\paragraph{Behaviour across masking rates.}
PM degrades sharply as $p_{\text{mask}}$ increases (CharFid drop of $0.12$–$0.26$ from $0.2 \to 0.8$), whereas EPC remains flat in the $0.98$–$1.00$ band. EPC's residual stream converts prediction errors into a compact, tunable correction channel.

\subsection{Alternative similarity metrics (ChrF and BERTScore)}
\begin{figure}[ht]
  \begin{adjustbox}{width=0.92\linewidth, center}
    \includegraphics{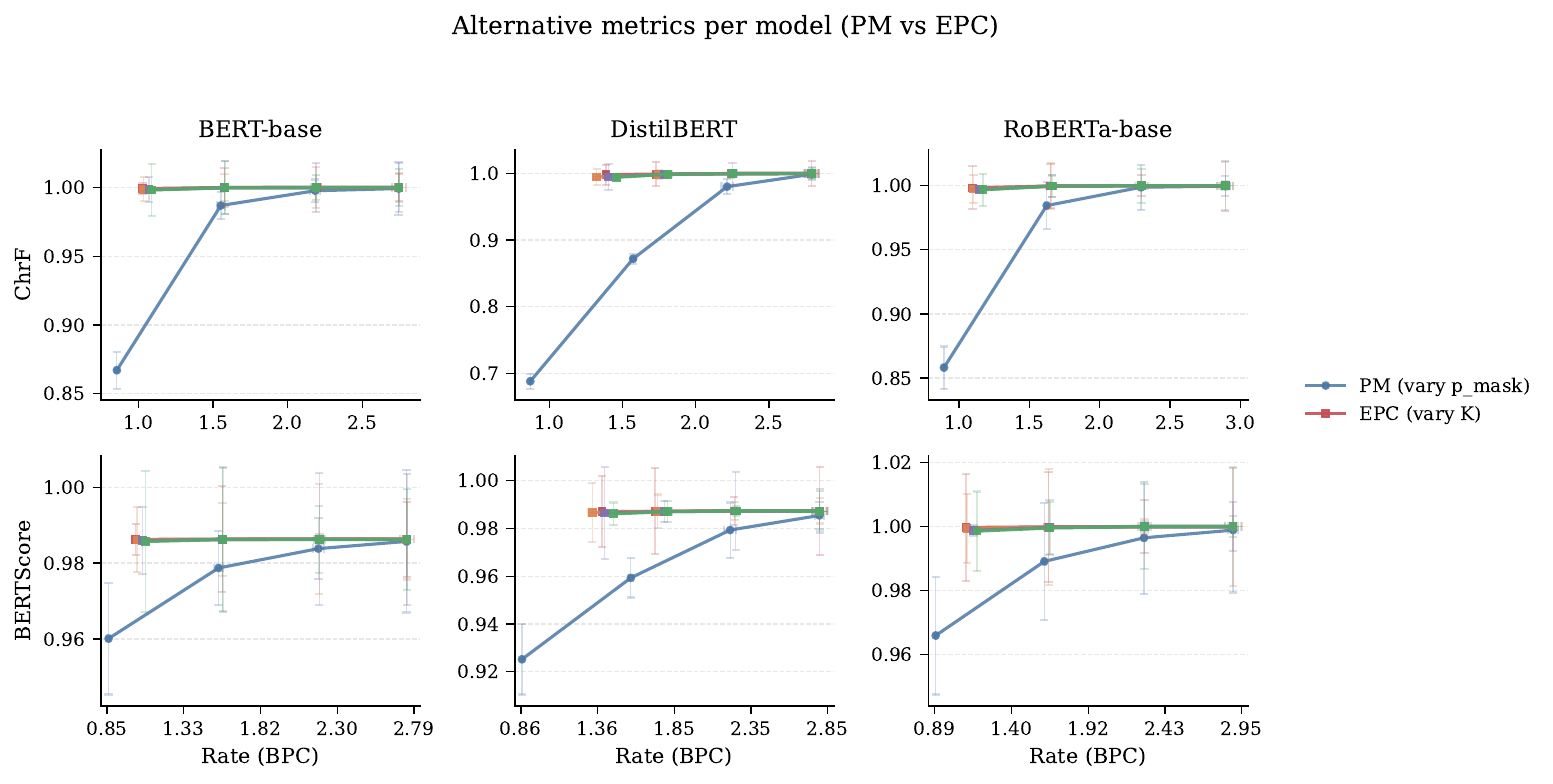}
  \end{adjustbox}
  \caption{\textbf{ChrF/BERTScore vs. BPC.} PM degrades as $p_{\text{mask}}$ grows; EPC stays saturated across rates, especially with RoBERTa.}
  \label{fig:rd-alt}
\end{figure}

\begin{figure}[ht]
  \begin{adjustbox}{width=0.92\linewidth, center}
    \includegraphics{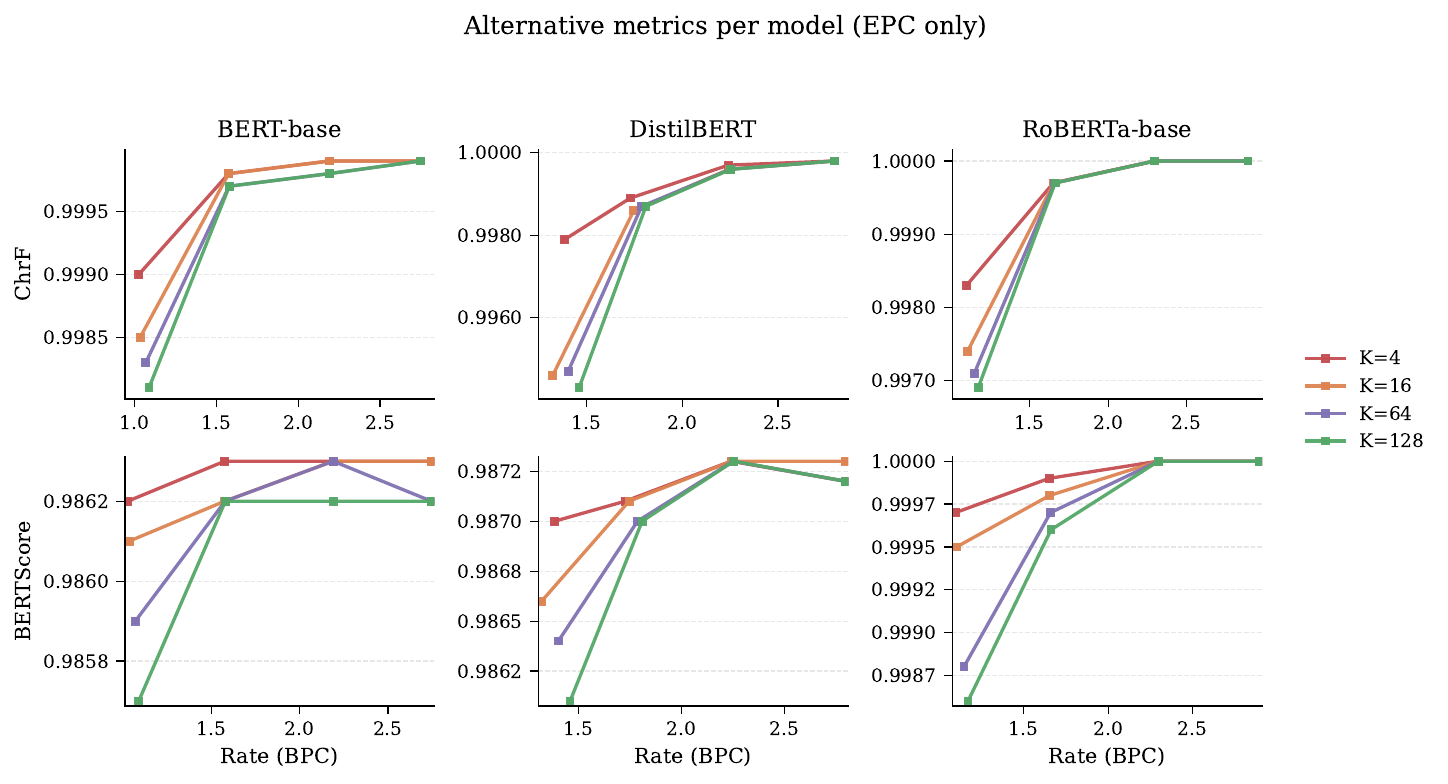}
  \end{adjustbox}
  \caption{\textbf{EPC across models.} RoBERTa delivers the strongest RD curve; BERT-base and DistilBERT are slightly behind but still markedly outperform PM.}
  \label{fig:rd-epc}
\end{figure}

As shown in Figure~\ref{fig:rd-alt}, the conclusions mirror those of CharFid: PM's performance declines with aggressive masking, while EPC remains near saturation. RoBERTa-EPC is effectively near-lossless (ChrF/BERTScore $\approx1.0$ at $p_{\text{mask}}\leq 0.4$). Figure~\ref{fig:rd-epc} isolates EPC performance to clarify cross-model differences.

\subsection{VQ+RE: a transform-coding point of comparison}
VQ+RE occupies a different RD regime, reaching low rates ($0.12$–$0.45$ BPC on BERT-base) but at substantially lower fidelity (CharFid $\approx0.65$–$0.79$). Standard deviations over seeds are larger (on RoBERTa-base, BPC sd spans $0.002$–$0.077$; CharFid sd spans $0.006$–$0.141$), reflecting the stochasticity of quantised latent training. Anchoring tokens slightly improves stability but increases the rate with modest fidelity gains.

\subsection{Ablations and knobs}
\paragraph*{EPC rank threshold $K$.} In our range $K\in\{4,16,64,128\}$, EPC's CharFid remains effectively constant while the rate moves slightly. This indicates that most ground-truth tokens are already within small ranks under the specialised MLM; $p_{\text{mask}}$ is the primary rate knob, with $K$ providing a fine-grained trade-off.
\paragraph*{Anchors in VQ+RE.} Introducing anchors reduces error cascades but linearly increases the rate. The anchor cost dominates the RD movement; the CharFid benefit is small relative to the added bits.

\subsection{Limitations}
Our research has several constraints. (i) \textbf{Model dependence and domain shift}. EPC relies on an MLM capable of acting as a decompressor; EPC's quality deteriorates when the evaluation domain differs from the MLM's pre-training or fine-tuning domain. EPC also uses a rank PMF for estimation, which is dependent on the corpus in use; therefore, calibration learned in one domain will not generalise to other domains. (ii) \textbf{Masking policy sensitivity}. Windowed equalisation reduces variability during the decoding process; however, extremely aggressive masking policies can cause localised error cascades when ground truth anchor tokens are sparse, particularly using weaker MLM models. (iii) \textbf{Coder approximations}. For our vocabulary fallback, we present two estimates of the arithmetic lower bound: an adaptive estimate and a practical rANS approximation that combines the negative mass $s$ into the merged mass of $\neg s$. The rANS approximation is slightly optimistic about the lower bound achievable with the complete CDF over the entire vocabulary space $V$. (iv) \textbf{Metrics}. CharFid/ChrF/BERTScore correlate well with the surface and semantic similarities of decoded outputs; they should not be considered as replacements for human evaluations. Factual and/or discourse errors may be under-penalised by these metrics. (v) \textbf{Compute and latency}. The cost of decompression offsets the rate reductions provided by EPC; i.e., the time required to perform iterative MLM infilling and possible additional refinement of the output exceeds the wall time of purely symbol-level compression algorithms. (vi) \textbf{Amortisation assumptions}. We amortise our BPC over $N$ copies of the same document; therefore, the effective rates per copy are inflated relative to a single user deployment where the MLM size remains constant. (vii) \textbf{VQ+RE scope}. The VQ+RE baseline was designed to operate at ultra-low rates; therefore, we have not explored alternative architectures or training methods. Accordingly, the fidelity of the VQ+RE baseline is likely to be better than what we reported.

\section{Conclusion}
We proposed a method for using masked language models (MLMs) as decompression methods for lossy compressed text, and we developed an approach called \textbf{Error-bounded predictive coding} (EPC). EPC sends rank-indexed residual data only when the top-ranked prediction from the model fails to meet the threshold. Using three different MLMs and across a broad range of compression ratios, we found that EPC achieves better performance on the rate-distortion curve compared to the baseline approach of \textbf{predictive masking} (PM): to achieve a CharFID score of approximately .98, EPC requires \textbf{52-63\% fewer bits} than PM, and at a compression ratio of approximately 1.6 BPC, EPC results in \textbf{+2.9 to +10.7} CharFID score points relative to PM. On our test dataset, EPC achieved nearly lossless quality using RoBERTa-base at approximately 1.1 BPC. We also evaluated a comparison case based on a transform coder (VQ+RE),, which can achieve compellingly low compression ratios, but at significantly lower fidelity than EPC. This suggests that EPC has advantages at higher compression ratios (or fidelities).

\paragraph{Future Outlook.} The masking rate and rank-threshold provide continuous rate distortion trade-offs. The potential future work includes: (i) joint training of the MLM and an EPC-aware objective function and learnable rank PMFs; (ii) adaptively setting the value of K and the budget for fallbacks to reduce error bounds; (iii) using a more sophisticated entropy coder to code the fallback stream; (iv) testing and evaluating the EPC system on multiple languages and outside of domain, including human evaluation; and (v) developing causal or streaming versions of EPC for use in real-time applications where latency matters. This line of development establishes EPC as a practical foundation for deploying high-quality lossy text compression systems.

\newpage
\bibliography{ref}
\pagebreak
\appendix

%%%%%%%%%%%%%%%%%%%%%%% DOCUMENT %%%%%%%%%%%%%%%%%%%%%%%%
\section{Detailed Codec Formalisms}\label{asec:formalism}

\subsection{Predictive Masking: Rate-Accurate Formulation}
Let $x_{1:N}$ be the token sequence. Let $\mathcal{M}\subset\{1,\dots,N\}$ be the masked indices, $|\mathcal{M}|=\lfloor p_{\text{mask}}N\rfloor$, and $\mathcal{S}=\{1,\dots,N\}\setminus\mathcal{M}$ the kept indices with fraction $p_{\text{keep}}=1-p_{\text{mask}}$. Denote model surprisal at $i$ by $s_i=-\log_2 q_\theta(x_i\mid X_{\setminus i})$.

\paragraph{Mask set selection.}
We use windowed equalisation. Partition $\{1,\dots,N\}$ into windows $\{W_w\}_w$; choose per-window thresholds $\tau_w$ such that
\begin{equation}
\label{eq:pm_window}
\begin{aligned}
\mathcal{M}&=\bigcup_w \bigl\{ i\in W_w : s_i\le \tau_w \bigr\},\\ |\mathcal{M}\cap W_w|&=\big\lfloor p_{\text{mask}}\cdot |W_w|\big\rfloor.
\end{aligned}
\end{equation}
We cap masked runs by enforcing a maximum run-length, which bounds local error cascades.

\paragraph{Payload.}
Positions are encoded with a succinct bitvector; the cost satisfies
\begin{equation}
\text{bits}_{\text{pos}}^{\text{PM}}\;\approx\; N \min\bigl\{H_2(p_{\text{keep}}),\ \mathcal{R}_{\text{RLE}}(p_{\text{keep}})\bigr\},
\label{eq:pm_pos}
\end{equation}
where $H_2(\cdot)$ is the binary entropy and $\mathcal{R}_{\text{RLE}}$ is the expected rate for run-length encoding. Kept tokens are entropy-coded in natural order with an auxiliary autoregressive coder $P_\psi$:
\begin{equation}
\text{bits}_{\text{tok}}^{\text{PM}}\;=\;\sum_{i\in\mathcal{S}} -\log_2 P_\psi\big(x_i\mid x_j: j\in\mathcal{S}, j<i\big).
\label{eq:pm_tok}
\end{equation}
Total payload bits for PM are $\text{bits}_{\text{PM}}=\text{bits}_{\text{pos}}^{\text{PM}}+\text{bits}_{\text{tok}}^{\text{PM}}$.

\paragraph{Reconstruction.}
We deterministically infill $\{x_i\}_{i\in\mathcal{M}}$ with $\arg\max$ under $q_\theta(\cdot\mid X_{\setminus \mathcal{M}})$, using the same specialisation from \hyperref[p:s1]{Stage~1}. The run-length cap guarantees at least one ground-truth token per window, which stabilises the local context during decoding.

\subsection{Error-Bounded Predictive Coding: Rank-Indexed Residuals}
Let $\mathcal{M}$ be selected as above. For each $i\in\mathcal{M}$, let $q_i(\cdot)=q_\theta(\cdot\mid X_{\setminus \mathcal{M}})$ be the MLM distribution and let $r_i\in\{1,2,\dots\}$ be the rank of the ground-truth token $x_i$ in the descending order of $q_i$’s probabilities. Fix a rank threshold $K\ge 2$.

\paragraph{Payload.}
Positions are coded as in \eqref{eq:pm_pos}. We introduce an override flag $z_i=\mathbf{1}[r_i>1]$ and encode it with a Bernoulli coder. With masked-set top-1 accuracy $p_1=\Pr(r_i=1)$,
\begin{equation}
\text{bits}_{\text{flag}}^{\text{EPC}}\;\approx\; |\mathcal{M}| H_2(1-p_1).
\label{eq:epc_flag}
\end{equation}
If $z_i=1$ and $2\le r_i\le K$, we transmit a rank index with code length $c_{\text{rank}}(r_i)$. If $r_i>K$, we fall back to the full token code of length $\ell(x_i)$. The correction stream cost is
\begin{equation}
\text{bits}_{\text{corr}}^{\text{EPC}}
=\sum_{i\in\mathcal{M}}\bigl(\mathbf{1}[2\le r_i\le K]\cdot c_{\text{rank}}(r_i)
+\mathbf{1}[r_i>K]\cdot \ell(x_i)\bigr).
\label{eq:epc_corr}
\end{equation}
Total payload bits for EPC are $\text{bits}_{\text{EPC}} \approx \text{bits}_{\text{pos}}^{\text{PM}} +\text{bits}_{\text{flag}}^{\text{EPC}} +\text{bits}_{\text{corr}}^{\text{EPC}}$.

\paragraph{Distortion control.}
If fallback is enabled for all $r_i>K$, EPC reconstructs the masked subset losslessly. In the lossy regime, we can disable the fallback or constrain it with a budget $\beta\in[0,1]$. The masked-set token error rate $D_{\text{masked}}$ is non-increasing in $K$ and $\beta$. This exposes two orthogonal controls: $p_{\text{mask}}$ trades positions versus modelling load, while $K$ and $\beta$ interpolate between a cheap residual stream and exact correction.

\paragraph{Reconstruction.}
At decode, we run the same specialised MLM to obtain the ranked list at each $i\in\mathcal{M}$, apply rank overrides where provided, and otherwise use $\arg\max$ as in PM.

\subsection{Vector Quantisation: Stabilised, Bounded-Error Formulation}
The goal is to train exactly what we deploy: keys/values are quantised during training, scheduled self-feeding removes train-test mismatch, and codebooks are learnt via EMA with utilisation regularisation.

\paragraph{Notation.}
Let $x_{1:T}$ be input tokens. The transformer has $L$ layers and $H$ heads. At layer $\ell$, token $t$ has hidden state $h_t^{(\ell)}\in\mathbb{R}^d$. For head $h$, queries, keys, and values are computed as $Q_t^{(\ell,h)} = h_t^{(\ell-1)} W_Q^{\ell,h}$, $K_u^{(\ell,h)} = h_u^{(\ell-1)} W_K^{\ell,h}$, $V_u^{(\ell,h)} = h_u^{(\ell-1)} W_V^{\ell,h}$, with $W_{\cdot}^{\ell,h}\in\mathbb{R}^{d\times d_h}$. Each head maintains two codebooks $\mathcal{C}_{K}^{\ell,h}=\{c_{K,j}^{\ell,h}\}_{j=1}^{K_K}$ and $\mathcal{C}_{V}^{\ell,h}=\{c_{V,j}^{\ell,h}\}_{j=1}^{K_V}$, where $c_{\cdot,j}^{\ell,h}\in\mathbb{R}^{d_h}$.

\paragraph{Quantise K/V at training time.}
Nearest-neighbour quantisation maps vectors to codebook entries:
\begin{equation}
\kappa_u^{(\ell,h)}=\arg\min_{j}\|K_u^{(\ell,h)}-c_{K,j}^{\ell,h}\|_2^2,\quad
\tilde{K}_u^{(\ell,h)}=c_{K,\kappa_u^{(\ell,h)}}^{\ell,h},
\end{equation}
and similarly for values $V$ to get $\tilde{V}_u^{(\ell,h)}$. Attention then uses these quantised vectors:
\begin{equation}
\begin{aligned}
A_{t,u}^{(\ell,h)}&=\mathrm{softmax}_u\left(\frac{Q_t^{(\ell,h)}\,\tilde{K}_u^{(\ell,h)\top}}{\sqrt{d_h}}\right),\\
o_t^{(\ell,h)}&=\sum_{u=1}^{T}A_{t,u}^{(\ell,h)}\tilde{V}_u^{(\ell,h)}.
\end{aligned}
\end{equation}
We apply the straight-through estimator (STE) for backpropagation.

\paragraph{Scheduled self-feeding.}
To remove exposure bias between training and testing, we replace teacher activations with their quantised counterparts with probability $\alpha_\tau$ at the training step $\tau$, using an inverse sigmoid schedule. At test time, $\alpha_\tau=1$.

\paragraph{EMA codebooks and commitment.}
Codebooks are updated using an exponential moving average (EMA) with decay $\rho$. We also added a commitment loss:
\begin{equation}
\mathcal{L}_{\text{commit}}=
\beta\sum_{u}\|\mathrm{sg}[K_u]-\tilde{K}_u\|_2^2
+\gamma\sum_{u}\|K_u-\mathrm{sg}[\tilde{K}_u]\|_2^2,
\end{equation}
and analogously for $V$, where $\mathrm{sg}[\cdot]$ is the stop-gradient operator.

\paragraph{Code usage regularisation.}
To avoid dead codes, we penalise the divergence of the empirical code usage distribution from a uniform prior.

\paragraph{Full training objective.}
The final loss combines the standard cross-entropy language modelling loss with VQ-specific terms:
\begin{equation}
\mathcal{L}
=\mathcal{L}_{\text{CE}}
+\mathcal{L}_{\text{commit}}
+\mathcal{L}_{\text{util}}^{K}+\mathcal{L}_{\text{util}}^{V}
+\mathcal{L}_{\text{sem}},
\end{equation}
where $\mathcal{L}_{\text{sem}}$ is a lightweight semantic consistency loss. Training uses the quantised $\tilde{K},\tilde{V}$ (blended by $\alpha_\tau$), EMA updates, and the STE.

\end{document}